\documentclass[letterpaper, 10 pt, conference]{ieeeconf}
\IEEEoverridecommandlockouts
                                                          
\usepackage{amsmath,amssymb,amsfonts}
\usepackage{graphicx}
\usepackage{textcomp}
\usepackage{xcolor}
\usepackage{hyperref}
\usepackage{amsmath}
\usepackage[noend]{algpseudocode}
\usepackage{comment}
\usepackage[normalem]{ulem}
\usepackage[framemethod=tikz]{mdframed}
\newtheorem{dfn}{Definition}
\newtheorem{asm}{Assumption}
\newtheorem{remark}{Remark}

\def\BibTeX{{\rm B\kern-.05em{\sc i\kern-.025em b}\kern-.08em
    T\kern-.1667em\lower.7ex\hbox{E}\kern-.125emX}}

\usepackage{graphicx}
\usepackage{psfrag}
\usepackage{pstool}
\usepackage{multirow}
\algnewcommand\algorithmicforeach{\textbf{for each}}
\algdef{S}[FOR]{ForEach}[1]{\algorithmicforeach\ #1\ \algorithmicdo}

\newcommand{\argmin}{\mathop{\mathrm{arg\,min}}}

\newcommand{\footref}[1]{%
    $^{\ref{#1}}$%
}

\usepackage{xspace}

\makeatletter

\DeclareRobustCommand\onedot{\futurelet\@let@token\@onedot}
\def\@onedot{\ifx\@let@token.\else.\null\fi\xspace}

\def\ie{\emph{i.e}\onedot}

\usepackage{algorithm,algpseudocode,setspace}

\usepackage{etoolbox}
\makeatletter
\patchcmd{\@makecaption}
  {\scshape}
  {}
  {}
  {}
\makeatletter
\patchcmd{\@makecaption}
  {\\}
  {.\ }
  {}
  {}
\makeatother
\def\tablename{Table}

\font\bfmath=cmmib10
\mathchardef\Gamma="7100
\mathchardef\Delta="7101
\mathchardef\Theta="7102
\mathchardef\Lambda="7103
\mathchardef\Xi="7104
\mathchardef\Pi="7105
\mathchardef\Sigma="7106
\mathchardef\Upsilon="7107
\mathchardef\Phi="7108
\mathchardef\Psi="7109
\mathchardef\Omega="710A

\mathchardef\alpha="710B
\mathchardef\beta="710C
\mathchardef\gamma="710D
\mathchardef\delta="710E
\mathchardef\epsilon="710F
\mathchardef\zeta="7110
\mathchardef\eta="7111
\mathchardef\theta="7112
\mathchardef\iota="7113
\mathchardef\kappa="7114
\mathchardef\lambda="7115
\mathchardef\mu="7116
\mathchardef\nu="7117
\mathchardef\xi="7118
\mathchardef\pi="7119
\mathchardef\rho="711A
\mathchardef\sigma="711B
\mathchardef\tau="711C
\mathchardef\upsilon="711D
\mathchardef\phi="711E
\mathchardef\chi="711F
\mathchardef\psi="7120
\mathchardef\omega="7121
\mathchardef\epsilon="7122

\mathchardef\varepsilon="7122
\mathchardef\vartheta="7123
\mathchardef\varpi="7124
\mathchardef\varrho="7125
\mathchardef\varsigma="7126
\mathchardef\varphi="7127
\mathchardef\imath="717B
\mathchardef\jmath="717C

\def\smallbfW{{\raise1.5pt\hbox{\mbox{\boldmath $_W$}}}}

\def\red{\color{red}}

\def\my4psfrag#1#2#3#4#5#6#7#8{
        \begin{figure}[htp]
        \begin{center}
            \begin{tabular}[h]{c c}
              {\leavevmode{\includegraphics[width=#1truecm]{#2.eps}}}
              &
              {\leavevmode{\includegraphics[width=#1truecm]{#3.eps}}} \\
              {\leavevmode{\includegraphics[width=#1truecm]{#4.eps}}}
              &
              {\leavevmode{\includegraphics[width=#1truecm]{#5.eps}}}
         \end{tabular}
           \vspace{#6}
           \caption{#7}
           \label{#8}
        \end{center}
        \end{figure}
}

\def\mydouble4psfrag#1#2#3#4#5#6#7#8{
        \begin{figure*}[htp]
        \begin{center}
            \begin{tabular}[h]{c c}
              {\leavevmode{\includegraphics[width=#1truecm]{#2.eps}}}
              &
              {\leavevmode{\includegraphics[width=#1truecm]{#3.eps}}} \\
              {\leavevmode{\includegraphics[width=#1truecm]{#4.eps}}}
              &
              {\leavevmode{\includegraphics[width=#1truecm]{#5.eps}}}
         \end{tabular}
           \vspace{#6}
           \caption{#7}
           \label{#8}
        \end{center}
        \end{figure*}
}

\makeatother
\usepackage[belowskip=-7pt,aboveskip=10pt,font={small}]{caption}
\usepackage[font={small}]{subcaption}

\AtBeginDocument{
    \setlength{\belowdisplayskip}{4pt} \setlength{\belowdisplayshortskip}{2pt}
    \setlength{\abovedisplayskip}{4pt} \setlength{\abovedisplayshortskip}{2pt}
}
\makeatletter
\let\origsection\section
\renewcommand\section{\@ifstar{\starsection}{\nostarsection}}

\newcommand\nostarsection[1]
{\sectionprelude\origsection{#1}\sectionpostlude}

\newcommand\starsection[1]
{\sectionprelude\origsection*{#1}\sectionpostlude}

\newcommand\sectionprelude{%
  \vspace{-1pt}
}

\newcommand\sectionpostlude{%
  \vspace{-1pt}
}

\let\origsubsection\subsection
\renewcommand\subsection{\@ifstar{\starsubsection}{\nostarsubsection}}

\newcommand\nostarsubsection[1]
{\subsectionprelude\origsubsection{#1}\subsectionpostlude}

\newcommand\starsubsection[1]
{\subsectionprelude\origsubsection*{#1}\subsectionpostlude}

\newcommand\subsectionprelude{%
  \vspace{-1pt}
}

\newcommand\subsectionpostlude{%
  \vspace{-1pt}
}

\let\origsubsubsection\subsubsection
\renewcommand\subsubsection{\@ifstar{\starsubsubsection}{\nostarsubsubsection}}

\newcommand\nostarsubsubsection[1]
{\subsubsectionprelude\origsubsubsection{#1}\subsubsectionpostlude}

\newcommand\starsubsubsection[1]
{\subsubsectionprelude\origsubsubsection*{#1}\subsubsectionpostlude}

\newcommand\subsubsectionprelude{%
  \vspace{1pt}
}

\newcommand\subsubsectionpostlude{%
  \vspace{-1pt}
}

\makeatletter

\makeatletter
\IEEEtriggercmd{\reset@font\normalfont\fontsize{7.5pt}{8.40pt}\selectfont}
\makeatother
\IEEEtriggeratref{1}
\begin{document}

\title{Latent Space Roadmap for Visual  Action Planning of Deformable and Rigid Object Manipulation
}




\author{ Martina Lippi*$^{1,2}$, Petra Poklukar*$^{1}$, Michael C. Welle*$^{1}$, Anastasiia Varava$^{1}$, \\Hang Yin$^{1}$, Alessandro Marino$^{3}$,   and Danica Kragic$^{1}$
\thanks{*These authors contributed equally (listed in alphabetical order).}
\thanks{ ${}^1$KTH Royal Institute of Technology Stockholm, Sweden}%
\thanks{ ${}^2$University of Salerno, Salerno, Italy}%
\thanks{ ${}^3$University of Cassino and Southern Lazio, Cassino, Italy}%
}

\maketitle

 \begin{abstract}
We present a framework for visual action planning of complex manipulation tasks with high-dimensional state spaces such as manipulation of deformable objects. Planning is performed in a low-dimensional latent state space that embeds images. 
We define and implement a Latent Space Roadmap (LSR) which is a graph-based structure that globally captures the latent system dynamics. Our framework consists of two main components: a Visual Foresight Module (VFM) that generates a visual plan as a sequence of images, and an Action Proposal Network (APN) that predicts the actions between them. We show the effectiveness of the method on a simulated box stacking task as well as a T-shirt folding task performed with a real robot. 
\end{abstract}

\section{Introduction and Related Work}
\label{sec:intro}
Designing efficient state representations for task and motion planning is a fundamental problem in robotics  studied for several decades~\cite{rosenschein1985formal, thrun1997probabilistic}. 
Traditional planning approaches rely on a comprehensive knowledge of the  state of the robot and the surrounding environment.  As an example, information about the robot hand and mobile base configurations as well as possible grasps is exploited in~\cite{lozano2014constraint} to accomplish sequential manipulation tasks. The space of all possible distributions over the robot state space, called belief space, is instead employed  in~\cite{kaelbling2013integrated} to tackle partially observable control problems. 

The two most important challenges in designing state representations for robotics are high dimensionality and complex dynamics of the state space. Sampling-based planning algorithms~\cite{Lav06} mitigate the first problem to a certain extent by randomly sampling the state space and hence avoiding representing it explicitly. However, when dealing with higher-dimensional spaces and more complex systems, such as highly deformable objects, these approaches become intractable~\cite{finn2017deep}. Moreover, analytical modeling of the states of these systems  and simulation of their dynamics in real time remains an open research problem~\cite{tang2018cloth}. 

\begin{figure}[htb!]
\begin{center}
\includegraphics[width=8cm]{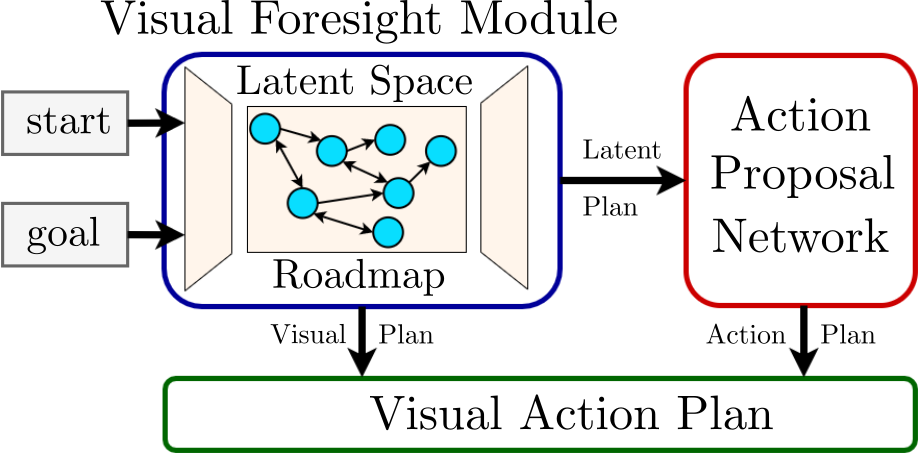}
\end{center}
\vspace{-10pt}
\caption{Overview of the proposed method.  The Visual Foresight Module (blue) takes the start and goal images and produces a visual plan from a latent plan found with the Latent Space Roadmap (cyan). The Action Proposal Network (red) proposes suitable actions to achieve the transitions between states in the visual plan. The final result is a \emph{visual action plan} (green) from start to goal containing actions to transition between consecutive states. }
\label{fig::overview_pipeline}
\end{figure}

For this reason, data-driven low-dimensional latent space representations for planning are receiving increasing attention as they make it possible to consider states that would otherwise be intractable. In particular, deep neural networks allow to implicitly represent complex state spaces and their dynamics 
thus enabling an automatic extraction of lower-dimensional state representations \cite{lesort2018state}. Unlike simpler methods, such as Principal Component Analysis (PCA)~\cite{wold1987principal}, deep neural networks also capture non-linear relations between features. Some of the most common approaches to learning compact representations in an unsupervised fashion are latent variable models such as Variational Autoencoders (VAEs) \cite{kingma2013auto, rezende2014stochasticvae2} or encoder-decoder based Generative Adversarial Networks (GANs) \cite{goodfellow2014generative, dumoulin2016adversarially}.  These models can learn low-dimensional state representations directly from images instead of a separate perception module. In this way, images can be used as input for planning algorithms to generate \emph{``visual plans"}~\cite{Ichter2019,nair2017combining}. 

Latent state representations, however, are not guaranteed to capture the \emph{global structure} and dynamics of the system, \ie to encode all the possible system states and respective feasible transitions. Furthermore, not all points in the latent space necessarily correspond to physically \emph{valid} states of the system, which makes it hard to plan by naively interpolating between start and goal states as shown in~Fig.~\ref{fig:sim1:super_results}. 
In addition, the transitions between the generated states might not be valid.
 
One way to address these shortcomings is to restrict the exploration of the latent space via imitation learning as presented  in~\cite{srinivas2018universal}, where a latent space Universal Planning Network (UPN) embeds differentiable planning policies and the process is learned in an end-to-end fashion. The authors then perform gradient descent to find optimal trajectories. 
 
A more common solution to mitigate the challenges of planning in latent spaces is to collect a large amount of training data that densely covers the state space and allows to infer dynamically valid transitions between states.  Following this approach, the authors in~\cite{Ichter2019} propose a framework for \emph{global search} in the latent space based on three components: \emph{i)} a latent state representation, \emph{ii)} a network that approximates the latent space dynamics, and \emph{iii)} a collision checking network. Motion planning is then performed directly in the latent space by an RRT-based algorithm. 
Similarly, a Deep Planning Network is proposed in~\cite{hafner2018learning} to perform continuous control tasks where a transition model, an observation model and a reward model in the latent space are learned and then exploited to maximize an expected reward function.
Following the trend of self-supervised learning, the manipulation of a deformable rope from an initial start state to a desired goal state is investigated in~\cite{wang2019learning}.  Building  upon~\cite{nair2017combining}, $500$ hours worth of data collection are used to learn the rope's inverse dynamics and then produce an \textit{understandable} visual foresight plan for the intermediate steps to deform the rope using a Context Conditional Causal InfoGAN ($C^3$IGAN).  

In this paper, we address the aforementioned challenges related to latent space representations by constructing a global roadmap in the latent space. Our Latent Space Roadmap (LSR) is a graph-based structure built in the latent space that both captures the \emph{global structure} of the state space and avoids sampling \emph{invalid} states. Our approach is data-efficient as we do not assume that the training dataset densely covers the state space neither accurately represents system dynamics. 
We instead consider a dataset consisting of \textit{pairs} of images and demonstrated actions connecting them, and then learn feasible transitions between states from this \textit{partial} data. This allows avoiding  full imitation for modeling as in the UPN framework~\cite{srinivas2018universal} as well as tackling tasks involving highly-deformable objects such as cloths.

More specifically, our method takes as input tuples consisting of an initial image, a successor image and properties of the action that occurred  between the states depicted in them. 
For example, in a box stacking task an action corresponds to moving one box, while in a T-shirt folding task it corresponds to making a fold (more details in Sec.~\ref{sec:exp}).  We deploy a VAE which we train with an augmented loss function that exploits the action information to enforce a more  favourable structure of the latent space.  A similar augmentation was explored in~\cite{rudolph2019structuring} but, in contrast to our work, uses an auto-encoder framework and requires class \textit{labels} which are not needed for LSR.  
Our method, visualised in Fig.~\ref{fig::overview_pipeline}, identifies the feasible transitions between regions containing similar states and generates a valid \emph{visual action plan} by sampling new valid states inside these regions. Our contributions can be summarized as follows: 

\begin{enumerate}
    \item  We define the Latent Space Roadmap that  enables generating a valid visual action plan. 
    While we use the VAE framework, our method can be applied to any other latent variable model with an encoder-decoder structure; 
    \item We augment the VAE loss function to encourage different states to be encoded further apart in the latent space and similar states to be encoded close by; 
    \item We experimentally evaluate our method on a simulated box stacking task as well as a real-world T-shirt folding task and quantitatively compare different  metrics in the loss augmentation ($L_1$, $L_2$, and $L_{\infty}$).  Complete details can be found on the website\footnote{\label{fn:website}\url{https://visual-action-planning.github.io/lsr/}}. 
\end{enumerate}

\section{Problem Statement and Notation} \label{sec:proplemdef}

{
The goal of visual action planning, also referred to as \emph{``visual planning and acting"} in~\cite{wang2019learning},  can be formulated as follows: given start and goal images, generate a path as a sequence of images representing intermediate states and compute dynamically valid actions between them. The problem is formalized in the following.

Let $\mathcal{I}$ be the state space of the system represented as images with fixed resolution and let $\mathcal{I}_{sys} \subset \mathcal{I}$ be the subset representing all the states of the system that are possible to reach while performing the task. A possible state $I \in \mathcal{I}_{sys}$ is called a \emph{valid state}. Let $\mathcal{U}$ be the set of possible control inputs or actions. 

\begin{dfn}
A \emph{visual action plan} consist of a \emph{visual plan} represented as a sequence of images $P_I = \{I_{start} = I_{0}, I_{1} ...,  I_N = I_{goal}\}$ where $ I_{start}$ and $I_{goal}$ are images representing the start and the goal states, and an \emph{action plan} represented as a sequence of actions $P_u = \{u_0, u_1, ..., u_{N-1}\}$ where $u_n \in \mathcal{U}$ generates a transition between consecutive states $I_n$ and $I_{n+1}$ for each $n\in\{0, ..., N-1\}$. 
\end{dfn}

To reduce the complexity of the problem we consider a lower-dimensional latent space $\mathcal{Z}$ encoding $\mathcal{I}$, and $\mathcal{Z}_{sys} \subset \mathcal{Z}$ encoding $\mathcal{I}_{sys}$.
Each image $I_n \in \mathcal{I}_{sys}$ can be encoded as a point $z_n \in \mathcal{Z}_{sys}$. Using $\mathcal{Z}_{sys}$, a visual plan can be computed in the latent space as  $P_z = \{z_{start} = z_0, z_1, ..., z_N = z_{goal}\}$ where $z_n \in \mathcal{Z}_{sys}$, and then decoded as a sequence of images. 
 
In order to obtain a valid visual plan, we study the structure of the space $\mathcal{Z}_{sys}$ which 
in general is  not path-connected. As we show in Sec.~\ref{sssec:vf-sim} and Fig.~\ref{fig:sim1:super_results},  
 linear interpolation between two valid states $z_1$ and $z_2$ in $\mathcal{Z}_{sys}$ may result in a path containing points from $\mathcal{Z} - \mathcal{Z}_{sys}$ that do not represent valid states of the system. To ensure a valid $P_z$, we therefore make an $\varepsilon$-validity assumption:
 
 \begin{asm}
 \label{eps-validity}
 Let $z \in \mathcal{Z}_{sys}$ be a valid latent state. Then there exists $\varepsilon > 0$ such that any other latent state $z'$ in the $\varepsilon-$neighborhood $N_{\varepsilon}(z)$ of $z$ is a valid latent state. 
\end{asm} 

This assumption, motivated by the continuity of the encoding of $\mathcal{I}$ into $\mathcal{Z}$, allows both taking into account the uncertainty induced by imprecisions in action execution and generating a valid visual plan. Each valid latent state $z$ in the visual plan can therefore be substituted by any other state $z'$ in the $\varepsilon-$neighborhood of $z$. To formalize this, we define an equivalence relation in $\mathcal{Z}_{sys}$
\begin{equation}\label{eq:equiv}
z \sim z' \iff ||z - z'||_d < \varepsilon,
\end{equation}
where the subscript $d \in \{1, 2, \infty\}$ denotes the metrics $L_1, L_2$ and $L_{\infty}$, respectively, and $\varepsilon$ a task-dependent parameter.

Consider a finite set of valid latent states \mbox{$\mathcal{R}_z = \{z_1, ..., z_M\} \subset \mathcal{Z}_{sys}$} induced by the set of valid input images  $\mathcal{R}_I = \{I_1, ..., I_M\} \subset \mathcal{I}_{sys}$. By Assumption~\ref{eps-validity} the union $\mathcal{R}_z^{\varepsilon}$ of $\varepsilon-$neighborhoods of the points in $\mathcal{R}_z$ consists of valid points:
\begin{equation}
\mathcal{R}_z^{\varepsilon} = \bigcup_{i = 1}^{M} N_{\varepsilon} (z_i) \subset \mathcal{Z}_{sys}. \label{def:regions}
\end{equation}

Assume that $\mathcal{R}_z^{\varepsilon}$ consists of $m$ path-connected components called \emph{valid regions} and denoted by $\{\mathcal{Z}_{sys}^i\}_{i=1}^{m}$. In general, if the points from $\mathcal{R}_z$ are sufficiently far away from each other, $m$ is larger than $1$. Note that each valid region is an equivalence class with respect to the equivalence relation \eqref{eq:equiv}. 
To connect them, we define a set of transitions between them:

\begin{dfn}
A \emph{transition function}  $f^{i, j}_z: \mathcal{Z}^{i}_{sys} \times \mathcal{U} \to \mathcal{Z}^{j}_{sys}$ maps any point $z \in \mathcal{Z}^{i}_{sys}$ to a class representative $z_{sys}^j \in \mathcal{Z}^{j}_{sys}$, where $i, j \in \{1, 2,..., m\}$ and $i \neq j$.
\end{dfn}

Given a set of valid regions $\mathcal{R}_{z}^\varepsilon$ in $\mathcal{Z}_{sys}$ and a set of transition functions connecting them we can approximate the global transitions of $\mathcal{Z}_{sys}$ as shown in Fig.~\ref{fig::visual_mani_plan}. To this end, we define a Latent Space Roadmap:

\begin{figure}[htb!]
\begin{center}
\includegraphics[width=0.33\textwidth]{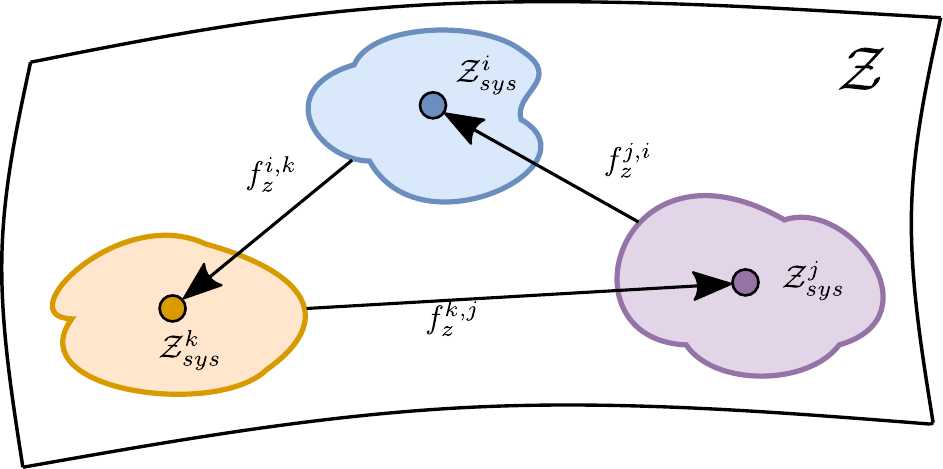}
\end{center}
\vspace{-15pt}
\caption{A visualisation of the structure of the latent state space showing valid regions $Z^i_{sys}$ and transition functions $f^{j,i}_z$ between them.
}
\label{fig::visual_mani_plan}
\end{figure}

\begin{dfn}
\label{def:lsr}
A Latent Space Roadmap is a directed graph $LSR = (\mathcal{V}_{LSR}, \mathcal{E}_{LSR})$ where each vertex $v_i \in \mathcal{V}_{LSR} \subset \mathcal{Z}_{sys}$ for $i \in \{1, 2,..., m\}$ is an equivalence class representative of the valid region $\mathcal{Z}^{i}_{sys}  \subset \mathcal{Z}_{sys}$, and an edge $e_{i, j} = (v_i, v_j) \in \mathcal{E}_{LSR}$ represents a transition function $f_z^{i, j}$ between the corresponding valid regions $\mathcal{Z}^{i}_{sys}$ and $\mathcal{Z}^{j}_{sys}$ for $i \neq j$. 
\end{dfn}

}

\section{An overview of our approach}
\subsection{Training Dataset}\label{ssec:dataset}
We consider a training dataset $\mathcal{T}_I$ consisting of generic tuples of the form $(I_1, I_2, \rho)$ where $I_1 \subset \mathcal{I}_{sys}$ is an image of the start state, $I_2 \subset \mathcal{I}_{sys}$ an image of the successor state, and $\rho$ a variable representing the action that took place between the two states. Here, an action is considered to be a \textit{single} transformation that produces any consecutive state $I_2$ different from the start state $I_1$, \ie, 
$\rho$ cannot be a composition of several transformations. 
On the contrary, we say that no action was performed if states $I_1$ and $I_2$ are variations of the same state, \ie, if the state $I_2$ can be obtained from $I_1$ with a small perturbation. The variable $\rho = (a, u)$ consists of a binary variable $a \in \{0, 1\}$ indicating whether or not an action occurred as well as a variable $u$ containing the task-dependent action-specific information which can be used to infer the transition functions $f_z^{i, j}$. For instance, an action in the box stacking example is illustrated in Fig.~\ref{fig:sim1:setup} where $u$ contains pick and place coordinates. If no action occurred, the right-hand side configuration in Fig.~\ref{fig:sim1:setup} would equal the one on the left-hand side with some small perturbations in the box positions. We call a tuple $(I_1, I_2, \rho = (1, u))$ an \textit{action pair} and  $(I_1, I_2, \rho = (0, u))$ a \textit{no-action pair}. When the specifics of the action $u$ are not relevant, we omit them from the tuple notation and simply write $(I_1, I_2, a)$. 
Finally, we denote by $\mathcal{T}_z$ the encoded training dataset $\mathcal{T}_I$ consisting of latent tuples $(z_1,z_2,\rho)$ obtained from the input tuples $(I_1,I_2,\rho) \in \mathcal{T}_I$ by encoding the inputs $I_1$ and $I_2$ in the latent space $\mathcal{Z}_{sys}$. 

\subsection{System Overview}
Our method consists of two main components depicted in Fig.~\ref{fig::overview_pipeline}. The first is the Visual Foresight Module (VFM) which is a trained VAE endowed with a Latent Space Roadmap (LSR). Given a start and goal state, the VFM produces a visual plan $P_I$ consisting of a sequence of images. The sequence $P_I$ is a decoded latent plan $P_z$ found in the VAE's latent space using the LSR.

The second component is the Action Proposal Network (APN) which takes a pair $(z_{i}, z_{i+1})$ of consecutive latent states from the latent plan $P_z$ produced by the VFM and proposes an action $u_i$ to achieve the desired transition \mbox{$f_z^{i, i+1}(z_i, u_i) = z_{i+1}$}. 

The two components combined produce a visual action plan that can be executed by any suitable framework. If open loop execution is not sufficient for the task, 
a re-planning step can be added after every action by substituting the start state with the current state and generating a new visual plan with  corresponding action plan.

\begin{remark}
Note that, although the tuples in the input dataset $\mathcal{T}_I$ contain only single actions $u$, our method is able to generate a sequence of actions $\{u_0, \dots, u_{N-1}\}$ to reach a goal state $I_N$ from a given start state $I_0$.
\end{remark}

\section{Visual Foresight Module (VFM)}
The Visual Foresight Module in Fig.~\ref{fig::overview_pipeline} has two building blocks that are trained in a sequential manner. Firstly, we train a VAE with an additional term in the loss function that affects the structure of the latent space. Once the VAE is  trained, we build our LSR in its latent space $\mathcal{Z}$ which identifies the valid regions $\mathcal{Z}^i_{sys}$. We present the details below.

\subsubsection{Latent state space}\label{sssec:vae}

Let $I \subset \mathcal{I}_{sys}$ be an input image, and let $z$ denote the unobserved latent variable and $p(z)$ the prior distribution. The VAE model~\cite{kingma2013auto, rezende2014stochasticvae2} consists of encoder and decoder neural networks that are jointly optimised to represent the parameters of the approximate posterior distribution $q(z|I)$ and the likelihood function $p(I|z)$, respectively. In particular, the VAE is trained to minimize 
\begin{align}
\mathcal{L}_{vae}(I) \!=\! E_{z \sim q(z|I)}[\log p(I|z)] + \beta \!\cdot\! D_{KL}\left(q(z|I) || p(z)\right) \label{eq:vaeloss}
\end{align} 
with respect to the parameters of the encoder and decoder neural networks. The first term influences the quality of the reconstructed samples, while the second term, called the KL divergence term, regulates the structure of the latent space. A better optimised KL term, achieved for example with a $\beta > 1$ \cite{higgins2016beta, burgess2018understanding}, results in a more compact latent space with points distributed according to the prior $p(z)$ but produces more blurry reconstructions. Therefore, the model needs to find a balance between the two opposing terms. 

Since our training data consists of tuples $(I_1, I_2, a)$, we compute $\mathcal{L}_{vae}$ for $I_1$ and $I_2$ separately and leverage the information contained in the  binary variable $a$ by minimizing an additional \emph{action} term
\begin{equation}\label{eq:acloss}
\mathcal{L}_{action}(I_1, I_2) \!=\! \begin{cases}
			\max(0, d_m - ||z_1-z_2||_d) & \text{if } a = 1\\
            ||z_1 - z_2||_d & \text{if } a = 0
		 \end{cases}
\end{equation}
where $z_1, z_2 \subset \mathcal{Z}_{sys}$ are the latent encodings of the input states $I_1, I_2 \subset \mathcal{I}_{sys}$, respectively, and the subscript $d$ denotes the metric as in \eqref{eq:equiv}. The hyperparameter $d_m$ introduced among the action pairs enforces different states to be encoded in separate parts of the latent space. 
The action term $\mathcal{L}_{action}$ naturally encourages the formulation of the valid regions  $\mathcal{Z}_{sys}^i$ in the latent space while maintaining the capability to generalise, \ie to sample novel valid states, inside each region $\mathcal{Z}_{sys}^i$.

The complete VAE loss term then equals
\begin{equation}\label{eq:loss}
\mathcal{L}(I_1, I_2) =  \frac{1}{2}( \mathcal{L}_{vae}(I_1) +\mathcal{L}_{vae}(I_2))   + \gamma \cdot \mathcal{L}_{action}(I_1, I_2)
\end{equation}
where the parameter $\gamma$ controls the influence of the distances among the latent codes on the structure of the latent space. 

\subsection{Latent Space Roadmap (LSR)}

The Latent Space Roadmap  is defined in \textit{Definition~\ref{def:lsr}} and built following the procedure summarised in Algorithm \ref{alg::fix_epsilon}. It is based on the idea that each node in the roadmap is associated with a valid region $\mathcal{Z}_{sys}^i$. Two nodes are connected by an edge if there exists an action pair $(I_1, I_2, \rho)$ in the training dataset $\mathcal{T}_I$ such that the transition $f_z^{1, 2}(z_1, u_1) = z_2$ is achieved in~$\mathcal{Z}_{sys}$. 

\begin{algorithm} \caption{LSR building}
\small
\def\negsp{\vspace{-5pt}}
\def\negup{\vspace{-7pt}}
\def\negin{\vspace{-3pt}}
\setstretch{1.2}
\begin{algorithmic}[section]
\Require Dataset $\mathcal{T}_z$, neighborhood size $\varepsilon$, metric $d$
\negin
\State Phase 1
\negsp
\begin{algorithmic}[1]
\State init graph $\mathcal{G} = (\mathcal{V}, \mathcal{E}) := (\{\}, \{\})$
\ForEach {($z_1,z_2, a) \in \mathcal{T}_Z $}
    \State $\mathcal{V} \gets $ create nodes $z_1,z_2$
    \If{$a=1$}
        \State $\mathcal{E} \gets $ create edge ($z_1,z_2$)
    \EndIf
\EndFor
\end{algorithmic}
\negup
\end{algorithmic}
\begin{algorithmic}[section]
\State Phase 2 
\negsp
\begin{algorithmic}[1]
\State $\mathcal{R}_{z}^{\varepsilon}:=\{\}$ 
\State $\mathcal{H} := \mathcal{V}$
\State $i := 1$
\While{$\mathcal{H} \neq \emptyset$} 
    \State randomly select $z \in \mathcal{H}$
    \State $\mathcal{W}^{i}_{sys} := \{z\}$ 
    
        \ForEach {$w \in \mathcal{W}^{i}_{sys}$}
            \State $\mathcal{W}^{i}_{sys} := \mathcal{W}^{i}_{sys} \cup \{w' \in \mathcal{H}:\|w-w'\|_d < \varepsilon \}$
        \EndFor
        \State $\mathcal{H} := \mathcal{H} \setminus \mathcal{W}^{i}_{sys}$
    \State $\mathcal{Z}^{i}_{sys} := \cup_{w \in \mathcal{W}^{i}_{sys}} N_\varepsilon(w) $
    \State $\mathcal{R}_{z}^{\varepsilon} := \mathcal{R}_{z}^{\varepsilon} \cup \{\mathcal{Z}^{i}_{sys}\}$ 
    \State $i := i + 1$
\EndWhile

\end{algorithmic}

\end{algorithmic}

\begin{algorithmic}[section]
\negup
    \State Phase 3
    \negsp
    \begin{algorithmic}[1]
        \State init graph $\text{LSR} = (\mathcal{V}_{LSR}, \mathcal{E}_{LSR}) := (\{\}, \{\})$
        \ForEach {$\mathcal{Z}^{i}_{sys} \in \mathcal{R}_{z}^{\varepsilon} $}
            \State $w^i_{sys} := \frac{1}{|\mathcal{W}^{i}_{sys}|} \sum_{w \in \mathcal{W}^{i}_{sys}}w$ 
            \State $z^i_{sys} := \text{argmin}_{z \in \mathcal{Z}_{sys}^i} ||z - w^i_{sys}||_d$
            \State $\mathcal{V}_{LSR} \gets $ create node $z^i_{sys}$
        \EndFor
        \negsp
        \ForEach {edge $ e = (v_1, v_2) \in \mathcal{E} $}
            \State find $\mathcal{Z}^{i}_{sys}, \mathcal{Z}^{j}_{sys}$ containing $v_1, v_2$, respectively 
            \State $\mathcal{E}_{LSR} \gets $ create edge ($z^{i}_{sys},z^{j}_{sys}$)
        \EndFor
    \end{algorithmic}
    \negsp
    \Return LSR
    
\end{algorithmic}
\label{alg::fix_epsilon}
\end{algorithm}

More specifically, the algorithm takes as an input the encoded training data $\mathcal{T}_z$, the parameter $\varepsilon$ defined in Assumption~\ref{eps-validity}  inducing the size of the valid regions $\mathcal{Z}_{sys}^i$, and the metric $d$ with respect to which we measure if a valid latent state is in the $\varepsilon-$neighbourhood of another valid state. Note that, as in the case of the VAE  (Sec.~\ref{sssec:vae}),  no action-specific information $u$ is used but solely the binary variable $a$ indicating the occurrence of an action.

Algorithm \ref{alg::fix_epsilon} consists of three phases. In Phase $1$, we build a \emph{reference} graph $\mathcal{G} = (\mathcal{V}, \mathcal{E})$  induced by $\mathcal{T}_z$ (lines $1.1 - 1.5$). Its vertices are all the latent states in $\mathcal{T}_z$ and edges exists only among the latent action pairs. It serves as a look-up graph to keep track of which areas in $\mathcal{Z}_{sys}$ have already been explored as well as to preserve the edges that later induce the transition functions $f_z^{i, j}$.

In Phase $2$, we identify the valid regions $\mathcal{Z}_{sys}^i \subset \mathcal{Z}_{sys}$. We start by randomly selecting a vertex from $\mathcal{V}$ and finding all the vertices from $\mathcal{V}$ that are in its $\varepsilon-$neighbourhood (lines $2.5 - 2.8$). 
The set $\mathcal{W}_{sys}^i$ of all the points found in this way necessarily belongs to the same connected component by Assumption~\ref{eps-validity}. However, we need to keep repeating this search for all the identified points (line $2.7$) as there might be more latent states in their respective $\varepsilon-$neighbourhoods. Once $\mathcal{W}_{sys}^i$ stops growing, the union of all $\varepsilon$-neighbourhoods of points in $\mathcal{W}_{sys}^i$ identifies the first connected component $\mathcal{Z}_{sys}^i$ (line $2.10$). We remove the set of allocated points $\mathcal{W}_{sys}^i$ from the reference vertex set $\mathcal{V}$ (line $2.9$) and continue identifying new valid regions until we considered all the points in $\mathcal{V}$ (line $2.4$). At the end of this phase we obtain the union of the valid regions $\mathcal{R}_z^\varepsilon = \{\mathcal{Z}_{sys}^i\}_i$.

In Phase $3$, we build the $\text{LSR}= (\mathcal{V}_{LSR}, \mathcal{E}_{LSR})$. We first compute the mean value $w_{sys}^i$ of all the points in each $\mathcal{W}_{sys}^i$ (line $3.3$). 
As the mean itself might not be contained in the corresponding path-connected component we find the class representative $z_{sys}^i \in \mathcal{Z}_{sys}^i$ that is the closest. The found representative then defines a node $v_i \in \mathcal{V}_{LSR}$ representing the valid region $\mathcal{Z}_{sys}^i$ (lines $3.4$ - $3.5$). Lastly, we use the set of edges $\mathcal{E}$ in the reference graph to infer the transition maps $f_z^{i, j}$ between the valid regions identified in Phase $2$. We create an edge in LSR if there exists an edge in $\mathcal{E}$ between two vertices in $\mathcal{V}$ that were allocated to different valid regions (lines $3.6 - 3.8$). 

The parameter $\epsilon$ is calculated as a weighted sum of the mean $\mu_0$ and the standard deviation $\sigma_0$ of the distances $\|z_1 - z_2\|_d$ among the no-action latent pairs $(z_1, z_2, a = 0)$:
\begin{equation}\label{eq:eps}
\epsilon= \mu_0  + w_\epsilon\cdot \sigma_0
\end{equation}
where $w_\epsilon$ is a scaling parameter that can be tuned for the task at hand. The rationale behind Eq.~\eqref{eq:eps} is that $\epsilon$ should be chosen such that  similar states, captured in the no-action pairs, belong to the same valid region, while states in the action pairs are allocated to different valid regions.

Using the LSR and the trained VAE-model, we can generate one or more visual plans from start to goal state. To this aim,  the states are first encoded in the latent space and the closest nodes in the LSR are found. Next, all shortest paths in the LSR~\cite{SciPyProceedings_11} between the identified nodes are retrieved. 
Finally, the class representatives of the nodes belonging to each shortest path compose the respective latent plan $P_z$, which is then decoded into the visual plan $P_I$. 

\section{Action Proposal Network (APN)} \label{sec:apn}
The Action Proposal Network is used to predict the specifics of an action $u_i$ that occurs between a latent pair $(z_i, z_{i+1})$ from a latent plan $P_z$ produced by the VFM.  We deploy a diamond-shaped multi layer perceptron and train it in a supervised fashion on the latent \emph{action} pairs 
$(z_1, z_2, \rho = (1, u))$ obtained from the enlarged dataset $\mathcal{T}_z$ as described below. The architecture details are reported in the code repository\footnote{\label{fn:git}\url{https://github.com/visual-action-planning/lsr-code}}. Since the network only depends on the action specifics $u$, it is easily adaptable to any task that fits the assumptions listed in Sec.~\ref{sec:proplemdef}. 

The training dataset for the APN is derived from $\mathcal{T}_I$ but preprocessed with the encoder of the trained VFM. In particular, for each training \emph{action} pair $(I_1, I_2, \rho = (1, u)) \in \mathcal{T}_I$ we first encode the inputs $I_1, I_2 \in \mathcal{I}_{sys}$ and obtain the parameters $\mu_{i}, \sigma_{i}$ of the approximate posterior distributions $q(z|I_i) =  N(\mu_i, \sigma_i)$, for $i = 1, 2$, given by the encoder network in the VFM. We then sample $2S$ novel points $z_1^s \sim q(z|I_1)$ and $z_2^s \sim q(z|I_2)$ for $s \in \{0, 1, \dots, S\}$. This procedure results in $S+1$ tuples $(\mu_{1}, \mu_{2}, \rho)$ and $(z_1^s, z^s_2, \rho), 0 \le s \le S$, where $\rho = (1, u)$ was omitted from the notation for simplicity. The set of all such low-dimensional tuples then forms a training dataset for the APN. 
\begin{remark}
It is worth remarking the two-fold benefit of this preprocessing step: not only does it reduce the dimensionality of the data but also enables enlarging it with novel points by factor $S+1$. 
\end{remark}

\section{Experiments}\label{sec:exp}
In this section, we evaluate the performance of the proposed approach both on a simulated box stacking task and on a real robotic hardware considering a T-shirt folding task. 
The purpose of the  simulation task is not to improve solutions for stacking boxes but to validate our approach in a quantitative and automatic manner where the ground truth is known. On the contrary, the T-shirt folding task evaluates the method in a complex scenario where highly  deformable objects are involved and the ground truth is unknown.

\subsection{Box stacking}\label{ssec:box}
\begin{figure}[htb!]
\begin{center}
\includegraphics[width=5.5cm]{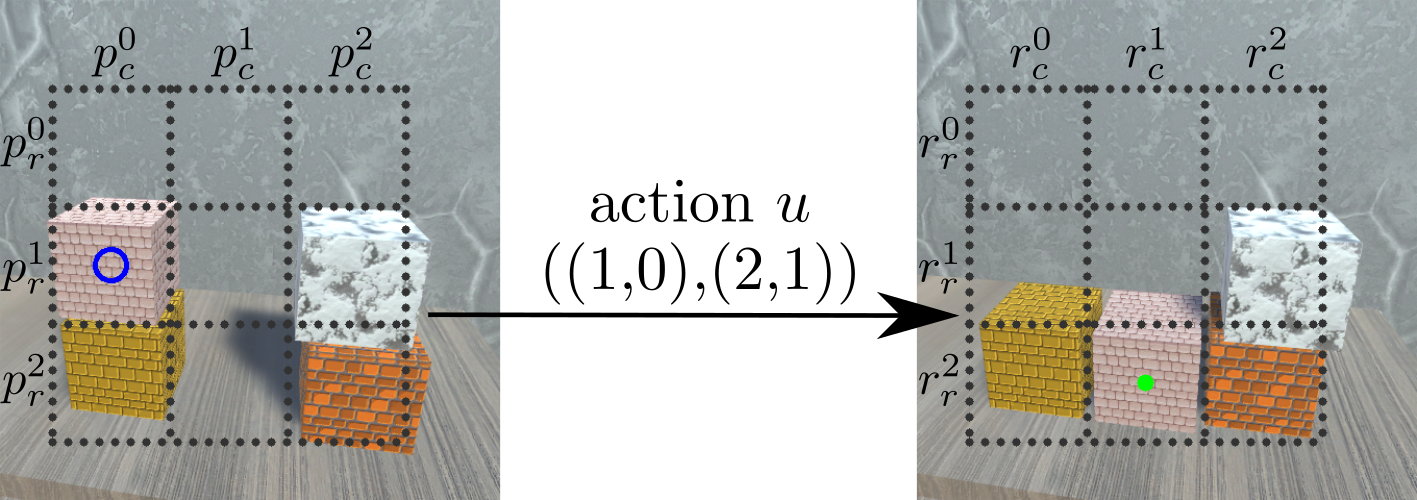}
\end{center}
\vspace{-12pt}
\caption{An example of an action $u = (p, r)$ in the box stacking task. The blue circle shows the picking location $p = (1, 0)$, and the green one the release position $r = (2, 1)$.}
\label{fig:sim1:setup}
\end{figure}

The simulation setup, shown in Fig.~\ref{fig:sim1:setup} and developed with the Unity engine~\cite{unitygameengine} 
is composed of four boxes with different textures that can be stacked in a $3 \times 3$ grid (dotted lines). A grid cell can be occupied by only one box at any time and a box can be moved according to the  \emph{stacking rules}: i) it can be picked only if there is no other box on top of it, ii) it can be released only on the ground or on top of another box inside the $3 \times 3$ grid. The action-specific information $u$, as shown in Fig.~\ref{fig:sim1:setup},  is a pair $u = (p, r)$ of pick $p$ and release $r$ coordinates in the grid  modelled by the row and column indices, \ie, $p = (p_r, p_c)$ with $p_r, p_c \in \{0, 1, 2\}$, and equivalently for \mbox{$r = (r_r, r_c)$}.

To have a diverse dataset with variation, the position of each box in a grid cell is generated by introducing $\sim 17\%$ noise along $x$ and $y$ axes, which is applied both when generating an action and a no-action pair. Each image is of dimension $256 \times 256 \times 3$. 
For the VFM, we deploy a VAE with a ResNet architecture~\cite{ResNet} for the encoder and decoder networks and a $64$-dimensional latent space. It is trained for $500$ epochs on a training dataset $\mathcal{T}_I$ composed of $5000$ tuples, $65\%$ of which are action pairs and $35\%$ no-action pairs. We train a \emph{baseline} VAE (VAE-b) without the action term in Eq.~\eqref{eq:loss}, \ie with $\gamma = 0$, and three \emph{action} VAEs (VAE-$L_1$, VAE-$L_2$, VAE-$L_\infty$) with the action term using
$d = 1,2,\infty$, respectively. Weights $\beta$ and $\gamma$ from Eqs.~\eqref{eq:vaeloss} and~\eqref{eq:loss} are increased over epochs following a scheduling procedure
starting from $\beta = 0$ and $\gamma = 1$. This encourages the models to first learn to reconstruct the input images and then gradually structure the latent space. The minimum distance $d_m$ in Eq.~\eqref{eq:acloss} is set to $20$, $5$, and $2.5$ in VAE-$L_1$, VAE-$L_2$ and VAE-$L_\infty$, respectively. These values are defined approximately as the average distance between the latent action pairs encoded by the VAE-b.

Similarly, we train four APNs (APN-b, APN-$L_1$, APN-$L_2$ and APN-$L_\infty$) on the latent training datasets $\mathcal{T}_z$ doubled with $S = 1$ using VAE-b, VAE-$L_1$, VAE-$L_2$ and VAE-$L_\infty$, respectively. The models were trained for $200$ epochs and we use the validation split (corresponding to $20\%$ of the data) to extract the best performing ones that are used in the experiments. The complete details about VFM and APN hyperparameters can be found in the configuration files in the code repository\footref{fn:git}. 

The designed task contains exactly $288$ different grid configurations, \ie, the specification of which box, if any, is contained in each cell. Given a pair of such grid configurations and the ground truth stacking rules, it is possible to analytically determine whether or not an action is allowed between them. This enables an automatic evaluation of the structure of the latent space $\mathcal{Z}_{sys}$, the quality of the  visual plan $P_I$ generated by the VFM as well as of the corresponding action plan ${P}_u$ predicted by the APN. We address these questions for all the action models and compare them to the baseline one in the experiments presented below.

\subsubsection{VAE latent space analysis}

In this section we discuss the influence of the action term~\eqref{eq:acloss} on the structure of the latent space $\mathcal{Z}_{sys}$.
Let each of the $288$ possible grid configuration represent a class. Note that each class contains multiple latent samples from the dataset $\mathcal{T}_z$ but their respective images look different because of the introduced positioning noise. Let $\bar z_c$ be the \emph{centroid} of the class $c$ defined as the mean point of the training latent samples $\mathcal{T}_z$ belonging to the class $c$. Let $d_{c,intra}^i$ be the \emph{intra-class} distance defined as the distance between a latent sample $z_i$ labeled with $c$ and the respective class centroid $\bar z_c$. Similarly, let $d_{i,inter}^j$ denote the \emph{inter-class} distance between the centroids $\bar z_i$ and $\bar z_j$ of classes $i$ and $j$. 

\begin{figure}[htb!]
\begin{center}
\includegraphics[width=8.6cm]{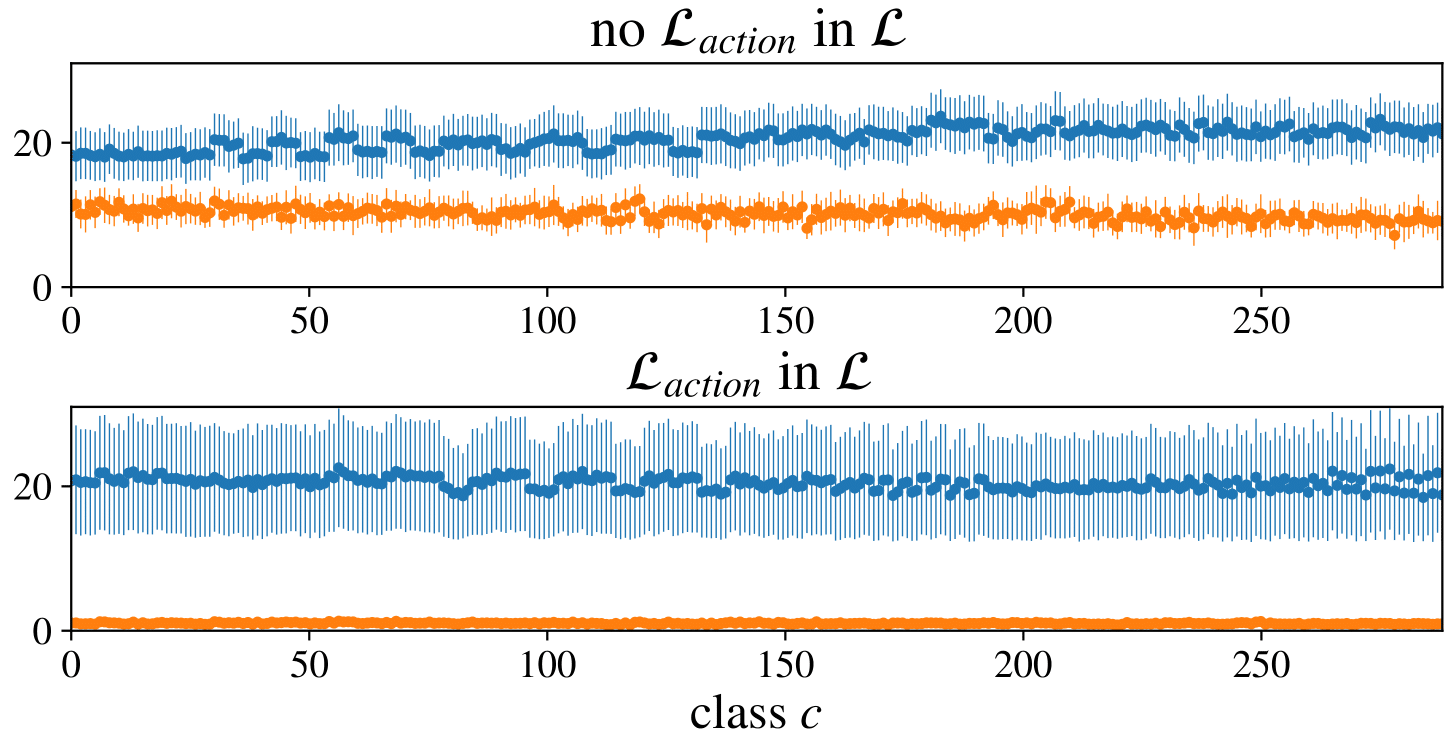}
\end{center}
\vspace{-15pt}
\caption{Mean values (bold points) and standard deviations (thin lines) of inter- (blue) and intra- (orange) distances for each class calculated using a VAE trained with (bottom) and without (top) action term.} 
\label{fig:sim1:dist}
\end{figure}

Fig.~\ref{fig:sim1:dist} reports the mean values (bold points) and the standard deviations (thin lines) of the inter-class (in blue) and intra-class (in orange) distances for each class $c\in\{1,...,288\}$. We compare the distances calculated using the latent training dataset $\mathcal{T}_z$ obtained from the baseline VAE (top) and the action VAEs (bottom). Due to the space constrains, we only report results obtained with metric $L_1$ but we observe the same behavior with $L_2$ and $L_{\infty}$.
In the case of baseline VAE, we observe similar intra-class and inter-class distances which 
implies that samples of different classes are encoded close together in  latent space and possible ambiguities may arise when planning in it. 
On the contrary, when using VAE-$L_1$ we observe that the inter- and intra-class distances approach the values $20$ and $0$, respectively, which are imposed with the action term~\eqref{eq:acloss} on the \emph{action} pairs and on not classes themselves. This means that, even when there exists no direct link between two samples of different classes and thus the action term for the pair is never activated, the VAE-$L_1$ is able to encode them such that the desired distances in the latent space are respected.

\begin{figure*}[htb!]
\begin{center}
\includegraphics[width=0.93\textwidth]{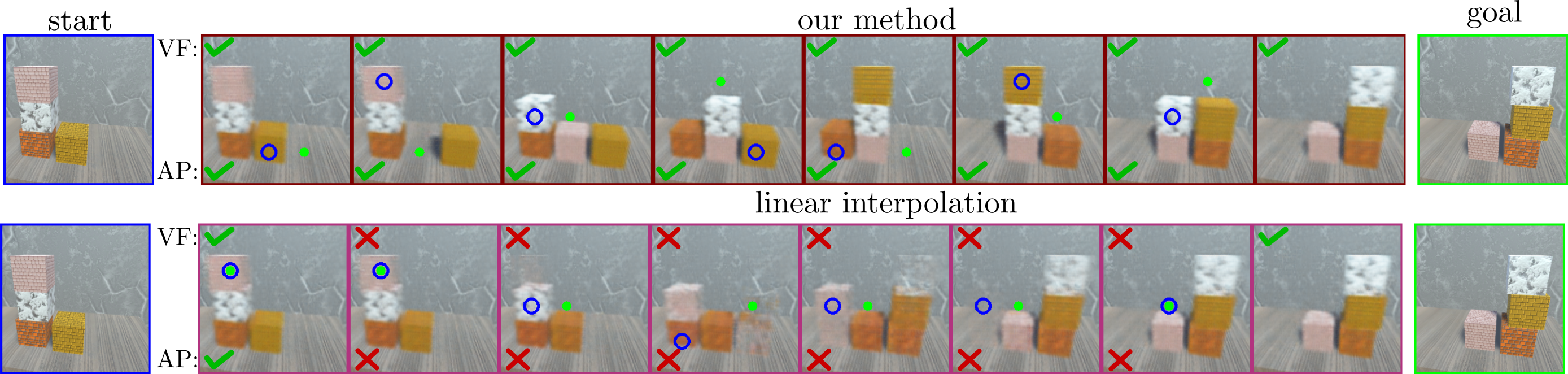}
\end{center}
\vspace{-10pt}
\caption{An example of a visual action plan from the start (left) to the goal state (right) for the box stacking task produced using our method (top) and a linear interpolation (bottom). Picking and releasing locations suggested by the APN are denoted with blue and green circles, respectively, while the outcome of the VFM (VF row) and the APN (AP row) are indicated with a green checkmark for success or a red X for failure. 
The APN succeeds using the path from our method and fails given the erroneous states of the linear interpolation. 
}
\label{fig:sim1:super_results}
\end{figure*}

In addition, we analyse the difference between the minimum inter-class distance and the maximum intra-class distance for each class. The higher the value the better separation of classes in the latent space is achieved. 
When the latent states are obtained using VAE-b we observe the difference to be always negative with an average value of $\approx -8.3$. On the other hand, when calculated on points encoded with VAE-$L_1$ it becomes non-negative for $286/288$ classes and its mean value increases to $\approx 0.78$. We therefore conclude that the action term results in a better structured latent space $\mathcal{Z}_{sys}$.

\subsubsection{LSR analysis}\label{sssec:vf-sim}
In this section we evaluate the quality of visual plans produced by our LSR in the latent space $\mathcal{Z}_{sys}$.

We consider three LSRs (LSR-$L_1$, LSR-$L_2$, LSR-$L_\infty$) that are built following Algorithm~\ref{alg::fix_epsilon} with the corresponding metrics using the latent training dataset $\mathcal{T}_z$ produced by either the baseline VAE or the action VAEs. The parameter $\varepsilon$  from Eq.~\eqref{eq:eps} is computed with a grid search on the weight $w_\epsilon \in \{-0.5, 0, 0.5,1\}$. Given a LSR, we evaluate its performance by measuring the quality of the visual plans found in it between $1000$ randomly selected start and goal states from an unseen test dataset of  $2500$ images. To this aim, a validity function\footref{fn:git} is defined that checks if a given visual action plan fulfills all the constraints determined by the stacking rules. 

In Table~\ref{tab:result_4_block_VFM} we show the results obtained on LSRs built with the training data from the baseline VAE (first row) and the action VAEs (last three rows). In particular, we report the percentage of cases when all the shortest paths in each LSR
are correct, when at least one of the proposed paths is correct, and the percentage of correct single transitions.

Firstly, we observe significantly worse performance of the LSRs when using the baseline VAE (first row) compared to using the action VAEs (bottom three rows). This indicates that VAE-b is not able to separate classes in $\mathcal{Z}_{sys}$ and we again conclude that the action term~\eqref{eq:acloss} needs to be included in the VAE loss function in  Eq.~\eqref{eq:loss} in order to obtain distinct valid regions~$\mathcal{Z}_{sys}^i$.

Secondly, we observe that among the action VAEs, LSR-$L_1$ outperforms the rest and is comparable with LSR-$L_2$, while LSR-$L_\infty$ reports the worst performances.
We hypothesise that this is because $L_1$ metric is calculated as the sum of the absolute differences between the individual coordinates and hence the points need to be evenly separated with respect to all dimensions. On the contrary, $L_\infty$ separates points based on only one dimension which leads to erroneous merges as two points might be far apart with respect to one dimension but very close with respect to the rest.

\begin{table}[]
\centering
\begin{tabular}{|c|l|l|l|l|l|}
\hline
Model  & All  & Any & Trans.  \\ \hline
VAE-b + LSR-$d,\,\forall d$   & $0$ \%        & $0$ \%       & $33.3$ \%    \\ \hline
VAE-$L_1$ + LSR-$L_1$  & {\boldmath${100}$} \%        & \boldmath{$100$} \%       & \boldmath{$100$} \%   \\ \hline
VAE-$L_2$ + LSR-$L_2$  & $99.9$ \%        & $99.9$ \%       & $99.9$ \%    \\ \hline
VAE-$L_\infty$ + LSR-$L_\infty$  & $12$ \%        & $8.2$ \%       & $53.2$ \%    \\ \hline
\end{tabular}
\vspace{-5pt}
\caption{Visual foresight results for box stacking case study comparing different metrics
(best results in bold). }
\label{tab:result_4_block_VFM}
\end{table}

\subsubsection{APN analysis}
We evaluate the accuracy of action predictions obtained by APN-b, APN-$L_1$, APN-$L_2$, and APN-$L_\infty$ on an unseen  test set consisting of $1491$ action pairs. As a proposed action can be binary classified as either true or false we calculate the percentage of the correct proposals for picking, releasing, as well as the percentage of pairs where both pick and release proposals are correct. All the models perform with $99\%$ or higher accuracy evaluated on $10$ different random seeds determining the training and validation sets\footref{fn:git}. This is because the box stacking task results in an $18$-class classification problem for action prediction which is simple enough to be learned from any of the VAEs.

Finally, we show the inadequacy of linear interpolation for the latent space planning. A linear visual path is obtained by  
uniformly sampling $n$ points along the line segment between given $z_{start}$ and $z_{goal}$
where $n$ equals the length of the shortest path retrieved from the LSR. An example of a linear visual path produced by VAE-$L_1$ is shown in the bottom row of Fig.~\ref{fig:sim1:super_results}. For each of the $1000$ start and goal states, decoding the linear paths with the VFM results in only failing transitions. In addition, invalid states are obtained where boxes of the same color are present multiple times and boxes exhibit \textit{invalid} states. On the contrary, a visual plan $P_I$ produced by LSR-$L_1$ using VAE-$L_1$ is  shown in the top row of Fig.~\ref{fig:sim1:super_results} and consists of only valid states and actions. 
 Moreover, the figure shows that the APN generalizes to the latent no-action pairs even though it is trained on the action pairs only.


\begin{figure*}[!ht]
\vspace{20pt}
\centering
\includegraphics[width=0.92\textwidth]{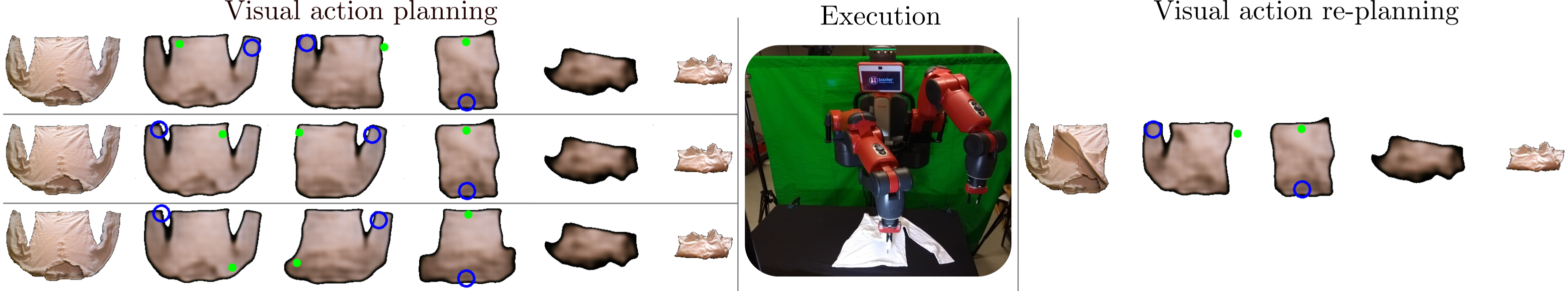}
 \vspace{-3pt}
\caption{Execution of the folding task with re-planning. On the left, a set of initial visual action plans reaching the goal state is proposed. After the first execution, only one viable visual action plan remains. }
\label{fig:folding:super_results_folding}
\end{figure*}

\subsection{T-shirt folding}
A Baxter robot, equipped with a Primesense RGB-D camera mounted on its torso, is used to fold a T-shirt in different ways as shown in 
Fig.~\ref{fig:folding:super_results_folding} and in the accompanying   video. All results are also reported in detail on the website\footref{fn:website}. 

For this task, a dataset $\mathcal{T}_I$ containing $1283$ pairs is collected. Each image has size $256\times256\times 3$, while the action specific information $u$ is defined as $u=(p,r,h)$  and is composed of picking coordinates $p=(p_r, p_c)$, releasing coordinates $r=(r_r, r_c)$ and picking height $h$. The values $p_r, p_c,r_r,r_c \in \{0,\dots, 255\}$ correspond to image coordinates, while $h \in \{0,1\}$ is either the height of the table or a 
value measured  from the RGB-D camera to  pick up only the top layer of the shirt. 

Note that the latter is a challenging task~\cite{seita2019deep} which is not in the scope of this work. The dataset $\mathcal{T}_I$ is collected by manually selecting pick and release points on images showing the current T-shirt configuration, and recording the corresponding action. No-action pairs are generated by slightly perturbing the cloth appearance, as shown in the video, which results in $37\%$ of no-action pairs in $\mathcal{T}_I$.

As shown in Fig.~\ref{fig:folding:super_results_folding}, we perform a re-planning step after each action execution to account for possible uncertainties. The current cloth state is then considered as a new start state and a new visual action plan is produced until the goal state ${I}_{goal}$ is reached  or the task is terminated. 
If multiple plans are generated, a human operator selects the one to  execute.

Compared to the box stacking task we use a larger version of the ResNet architecture for the VFM but keep the $64$-dimensional latent space. Following the model notation from Sec.~\ref{ssec:box}, we train a baseline VAE which we use to determine the minimum distance $d_m$ used in the action term~\eqref{eq:acloss} for the action VAEs. For the shirt folding task, these are set to $13.5,\, 3.5 $ and $2$ for VAE-$L_1$, VAE-$L_2$ and VAE-$L_\infty$, respectively. The APN models are trained using the same architecture as in the box stacking task and on training datasets enlarged with $S = 1$. Hyperparameters are similar to the box stacking experiment and can be found in the code repository\footref{fn:git}.

\subsubsection{APN Analysis}
We evaluate the performance of the APN models on $5$ random seeds on a test split consisting of $104$ action pairs. For each seed we reshuffle all the collected data and create new training, validation and test splits. The action coordinates $p$ and $r$ are first scaled to the interval $[0, 1]$, and then standardised with respect to the mean and the standard deviation of the training split.  

Table~\ref{tab:sim1:apn_results_folding} reports  mean and standard deviation of the Mean Squared Error calculated across the different random seeds. We separately measure the error obtained on picking predictions, releasing predictions, and the total error on the predictions of the whole action $u = (p, r, h)$. We observe a higher error when using VAE-b which again indicates that the latent space lacks structure if the action term~\eqref{eq:acloss} is excluded from the loss function. The best performance is achieved by APN-$L_1$ which corroborates the discussion from Sec.~\ref{ssec:box} about the influence of $L_1$ metric on the latent space.
\begin{table}[htb!]
\centering
\setlength{\textfloatsep}{0cm}
\setlength{\abovecaptionskip}{0cm} 
\setlength{\intextsep}{10pt plus 2pt minus 2pt}
\begin{tabular}{|c|l|l|l|}
\hline
Model & Pick & Release  & Total \\ \hline
APN-b  & $2.12 \pm 0.34$ & $2.21 \pm 0.13$   & $4.47 \pm 0.39$  \\ \hline
APN-$L_1$ & {\boldmath $1.86 \pm 0.11$} & {\boldmath $2.03 \pm 0.07$} & {\boldmath$3.96 \pm 0.16$}  \\ \hline
APN-$L_2$ & $2.06 \pm 0.14$ & $2.05 \pm 0.07$ & $4.22 \pm 0.2$     \\ \hline
APN-$L_\infty$ & $1.98 \pm 0.13$ & $2.16 \pm 0.1$     &  $4.3 \pm 0.16$  \\ \hline
\end{tabular}
\vspace{5pt}
\caption{The error of action predictions obtained in the folding task on APN models with different metrics (best results in bold).}
\label{tab:sim1:apn_results_folding}
\end{table}
\subsubsection{Execution Results}
The performance of the entire system cannot be evaluated in an automatic manner as in the box stacking task. We therefore choose five novel goal configurations and perform the folding task five times per configuration on each framework F-$L_d$ that uses VAE-$L_d$, APN-$L_d$, and LSR-$L_d$ with $d = 1, 2, \infty$. 
Weights $w_\varepsilon$ are experimentally set to $0.8$, $0.2$, and $-0.3$, respectively. 
In order to remove outliers present in the real data, a final pruning step is added to Algorithm~\ref{alg::fix_epsilon} which removes nodes from the $\mathcal{V}_{LSR}$ that contain less than $6$ training samples.


The results are shown in Table \ref{tab:folding:results_folding}, while all execution videos, including the respective visual action plans, are available on the website\footref{fn:website}. We report the total system success rate with re-planning, 
the percentage of correct single transitions, 
and the success of any  visual plan and  action plan from start to goal. 
Framework F-$L_1$ finds at least one visual action plan that makes the correct prediction, however, the execution of the action is not perfect. We therefore observe a lower overall system performance as the re-planning can result in a premature termination. 
Similar to the box-stacking task, results hint that F-$L_1$ is more suitable for executing the folding task while F-$L_\infty$ performs worst. 
\begin{table}[htb!]
\centering
\begin{tabular}{|c|l|l|l|l|}
\hline
          Framework     & Syst. & Trans.  & VFM & APN    \\ \hline
F-$L_1$  &     {\boldmath$80 \%$}   &  {\boldmath$90 \%$}      &  {\boldmath$100 \%$}       &      {\boldmath$100 \%$}                  \\ \hline
 F-$L_2$  &      $40 \%$      &  $77 \%$      &    $60 \%$           &       $60 \%$         \\ \hline
F-$L_\infty$&     $24 \%$    &   $44 \%$    &       $56 \%$       &        $36 \%$         \\ \hline

\end{tabular}
\vspace{-5pt}
\caption{Results (best in bold) for executing visual action plans on $5$ folding tasks (each repeated $5$ times). Different metrics are compared.  }
\label{tab:folding:results_folding}
\end{table}

Finally, a re-planning example is shown in Fig.~\ref{fig:folding:super_results_folding} where  a subset of  the proposed visual action plans is shown (left). As the goal configuration does not allude to how the sleeves are to be folded, the LSR suggests all paths it identifies. After the first execution, the re-planning  (right) generates in a single plan that leads from start to goal state.


\section{Conclusions and Future Work}

In this work, we addressed the problem of visual action planning. We proposed to build a Latent Space Roadmap which is a graph-based structure in a low-dimensional latent space capturing the  latent transition dynamics in a data-efficient manner.
Our method consists of a Visual Foresight Module, generating a visual plan from given start and goal states, and an Action Proposal Network, predicting the corresponding action plan.
 We showed the effectiveness of our method on a simulated box stacking task as well as a T-shirt folding task, requiring deformable object manipulation and performed with a real robot. 
As future work, we plan to extend the scope of the LSR to more domains such as Reinforcement Learning.


\bibliographystyle{IEEEtran}

\bibliography{bibliography}



\newpage

\end{document}